\documentclass[a4paper]{article}

\usepackage{INTERSPEECH2015}

\usepackage{graphicx}
\usepackage{amssymb,amsmath,bm}
\usepackage{textcomp,multirow}

\usepackage{comment}
\usepackage{bm}
\usepackage{hyperref}

\ninept

\title{Unsupervised Domain Discovery using Latent Dirichlet Allocation for Acoustic Modelling in Speech Recognition}

\makeatletter
\def\name#1{\gdef\@name{#1\\}}
\makeatother \name{\em Mortaza Doulaty, Oscar Saz, Thomas Hain 
}

\address{Speech and Hearing Group, University of Sheffield, Sheffield, UK \\
	{\small \tt \{mortaza.doulaty, o.saztorralba, t.hain\}@sheffield.ac.uk}
}

\begin{document}

  \maketitle
  \begin{abstract}

Speech recognition systems are often highly domain dependent, a fact widely reported in the literature.
However the concept of domain is complex and not bound to clear criteria. Hence it is often not evident
if data should be considered to be out-of-domain. While both acoustic and language models can be
domain specific, work in this paper concentrates on acoustic modelling. We present a novel
method to perform unsupervised discovery of domains using Latent Dirichlet Allocation (LDA) modelling. Here a set of hidden domains is assumed to exist in the data, whereby each audio segment can be considered to be a weighted mixture of
domain properties. The classification of audio segments into domains allows the creation of domain specific acoustic models for automatic speech recognition. Experiments are conducted on a dataset of diverse speech data covering speech from radio and TV broadcasts, telephone conversations, meetings, lectures and read speech, with a joint training set of 60 hours and a test set of 6 hours. Maximum A Posteriori (MAP) adaptation to LDA based domains
was shown to yield relative Word Error Rate (WER) improvements of up to 16\% relative, compared to pooled training,
and up to 10\%,  compared with models adapted with human-labelled prior domain knowledge.

  \end{abstract}

\vspace{-2mm}
  \section{Introduction}
\vspace{-2mm}
Recently, new applications and domains 
are becoming the target of research in Automatic Speech Recognition (ASR), as the existing systems increase their accuracy. 
This has opened the issue on how to scale up existing systems when new domains are incorporated as target data, for instance ``found data'', such as media and 
historical audio archives. In this situation, training acoustic models for an unknown domain, like different YouTube recordings, can 
be infeasible if the origin of the target speech can not be properly assessed, and the loss of 
accuracy is large due to wrong modelling decisions.

Well--tailored single domain systems, where training data that properly matches the target recognition data is available, are mostly used in current speech recognisers.
These domain dependent models have been usually trained via Maximum Likelihood (ML) if a sufficiently large amount of domain data existed or using adaptation techniques such as Maximum A Posteriori \cite{gauvain1994maximum}, Maximum Likelihood Linear Regression (MLLR) \cite{leggetter1995maximum} or  Cluster Adaptive Training \cite{gales1998cluster}. For more recent Deep Neural Network (DNN)--based systems, domain adaptation is also possible with linear transformations, conservative training and subspace methods \cite{deng2015} with frameworks such as Multi--Level Adaptive Networks (MLAN) \cite{bell2013multi} or Deep Maxout Networks (DMN) \cite{miao2013}.

An important issue when dealing with highly diverse speech data is the difficulty to appropriately categorise every speech input within a particular domain, especially the case with newly discovered data. Even when domain categories
have been given manually by humans, this may be inaccurate or there may be hidden characteristics in the audio that can further subdivide these categories or cross across several of the predefined domains.
Developing the ability of discovering these new and hidden acoustic domains would greatly enhance the possibility of using well--targeted specific domain models in ASR.
However, as most speech recognition tasks assume a single domain or well differentiated domains, the task of unsupervised discovery of acoustic domains in speech data has been of less interest so far. This paper proposes to open new areas for research in multi--domain ASR by treating speech data as a set of documents where latent domains exist and can be discovered
using Latent Dirichlet Allocation (LDA) models.

LDA is an statistical approach to discover latent topics in a collection of documents in an unsupervised manner \cite{blei2003latent}. It is mostly used in Natural Language Processing (NLP) for the categorisation of text documents, but it has been used for audio and image processing as well. In audio tasks, LDA has been used for classifying unstructured audio files into onomatopoeic and semantic descriptions with successful results \cite{kim2009audio,kim2009acoustic}. Building on this knowledge, this work proposes to use LDA for domain adaptation in ASR tasks.


This paper is organised as follows: Section \ref{sec:lda} will give an overview of LDA modelling in its original proposal for topic modelling. Then, Section \ref{sec:domain} will describe the proposed use of LDA models for unsupervised domain discovery in speech data. Section \ref{sec:setup} will present the experimental setup used for multi--domain speech recognition, with
Section \ref{sec:results} detailing the obtained results. Section \ref{sec:conclusion} gives the conclusions to this work.

\vspace{-2mm}
  \section{Latent Dirichlet Allocation}
  \label{sec:lda}
  \vspace{-2mm}
Latent Dirichlet Allocation (LDA) \cite{blei2003latent} is an unsupervised probabilistic generative model for collections of discrete data. It aims to describe how every item within the collection is generated, assuming that there are a set of hidden topics and that each item is modelled as a finite mixture over those topics. Also, an infinite mixture over an underlying set of topic probabilities is used to model each topic \cite{blei2003latent}. LDA is mostly used for topic modelling of text corpora, however, the model can be applied to other tasks, such as object categorisation and localisation in image processing \cite{sivic2005discovering}, automatic harmonic analysis in music processing \cite{hu2009probabilistic} or acoustic information retrieval in unstructured audio analysis \cite{kim2009acoustic}.

In the context of text corpora, a dataset is defined as a collection of documents and each document is a collection of words. Given a vocabulary of size $V$, each word is represented by a $V$--dimensional binary vector. It is assumed that the documents are generated using the following generative process:
\begin{enumerate}
	\item For each document $d_m, m \in \{1 ... M\}$, choose a $K$--dimensional topic weight vector $\theta_m$ from the Dirichlet distribution with scaling parameter $\alpha$: $p(\theta_{m}|\alpha)=Dir(\alpha) $
	\item For each word $w_n, n \in \{1 ... N\}$ in document $d_m$ 
	\begin{enumerate}
		\item Draw a topic $z_n \in \{1 ... K \}$ from the multinomial distribution $p(z_n=k|\theta_m)$ 
		\item Given the topic, draw a word from $p(w_n | z_n, \beta)$, where $\beta$ is a $V \times K$ matrix and \\$\beta_{ij}=p(w_n = i | z_n = j, \beta)$
	\end{enumerate}
\end{enumerate}

Other assumptions include the bag--of--words property of the documents and the fixed and known dimensionality of the Dirichlet distribution $K$ (and thus the dimensionality of the topic variable $z$)

\begin{figure}
	\centering
	\includegraphics[width=6cm]{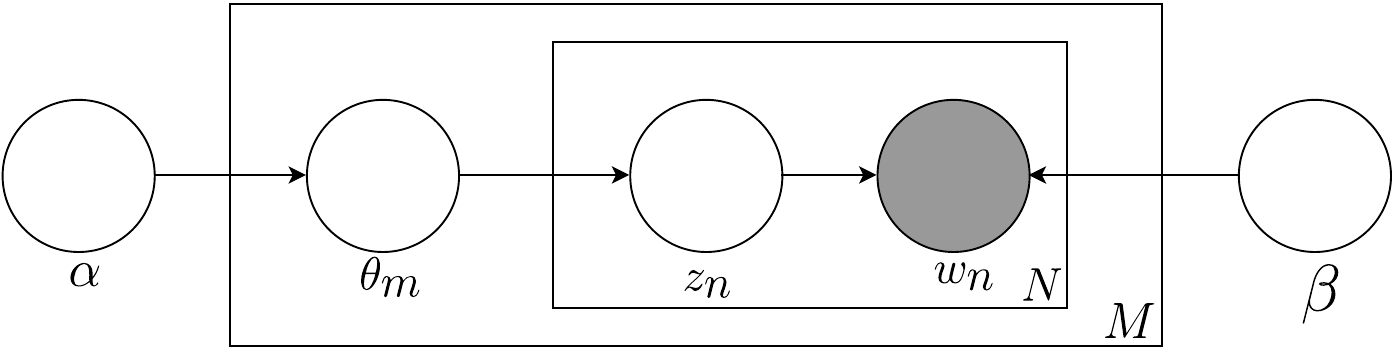}
	\caption{Graphical model representation of LDA}
	\label{fig:lda-graphical-model}
\end{figure}

The graphical representation of LDA model is shown at Figure \ref{fig:lda-graphical-model}, a three level hierarchical Bayesian model. In this model, the only observed variable is $w$ and the rest are all latent. $\alpha$ and $\beta$ are corpus level parameters, $\theta_m$ are document level variables and $z_n$, $w_n$ are word level variables. The generative process is described formally as:
\vspace{-1mm}
\begin{equation}
p(\theta, \mathbf{z}, \mathbf{w} | \alpha, \beta) 
= p(\theta | \alpha) \prod_{n=1}^{N}p(z_n | \theta) p(w_n|z_n,\beta)
\end{equation}
\vspace{-1mm}
The posterior distribution of the latent topic variables given the words and $\alpha$ and $\beta$ parameters is:
\vspace{-1mm}
\begin{equation}
	\label{eq:posterir}
	p(\theta, \mathbf{z} | \mathbf{w}, \alpha, \beta) = 
	\frac{p(\theta, \mathbf{z}, \mathbf{w} | \alpha, \beta)}{p(\mathbf{w} | \alpha, \beta)}
\end{equation}
\vspace{-1mm}
Computing $p(\mathbf{w} | \alpha, \beta)$ requires some intractable integrals. A reasonable approximate can be acquired using variational approximation which is shown to work reasonably well in various applications \cite{blei2003latent}. The approximated posterior distribution is:
\vspace{-1mm}
\begin{equation}
\label{eq:approx-posterior}
q(\theta, \mathbf{z} | \gamma, \phi) = q (\theta | \gamma) \prod_{n=1}^{N}q(z_n | \phi_n)
\end{equation}
\vspace{-1mm}
where $\gamma$ is the Dirichlet parameter that determine $\theta$ and $\phi$ is the parameter for the multinomial that generates the topics. 

Training tries to minimise the Kullback--Leiber divergence (KLD) \cite{kullback1951information} between the real and the approximated joint probabilities (equations \ref{eq:posterir} and \ref{eq:approx-posterior}) \cite{blei2003latent}:
\vspace{-1mm}
\begin{equation}
\underset{\gamma, \phi}{argmin} 
\; KLD \big(
q(\theta, \mathbf{z} | \gamma, \phi)
	 \; || \; 
p(\theta, \mathbf{z} | \mathbf{w}, \alpha, \beta)
\big)
\end{equation}
\vspace{-1mm}
Other training methods based on Markov--Chain Monte-Carlo is also proposed, like Gibbs sampling method \cite{griffiths2004finding}.

\section{Unsupervised Domain Discovery}
\label{sec:domain}
The proposed technique uses an LDA model to discover hidden and latent acoustic domains in multi--domain speech data.
Since LDA is for collections of discrete data (such as text corpora) \cite{blei2003latent}, 
every speech segment of length $T$ frames, $\mathbf{x}=\{\mathbf{x}_1,...,\mathbf{x}_t,...,\mathbf{x}_T\}$, is represented as a set of discrete symbols to support modelling within this framework. For that purpose, the $n$--dimensional audio frames, $\mathbf{x}_t\in\mathbb{R}^n$, are quantised into a dictionary of $V$ acoustic ``words'', $\bar{\mathbf{x}}_t\in\{1...V\}$ \cite{kim2009audio}. First a Gaussian Mixture Model (GMM) is trained using Expectation Maximisation (EM) and mix--up procedure to reach the desired codebook size $V$ (enforcing the co--variance matrix to be identity, equivalent to LBG--VQ \cite{gersho1992vector}). Then the means of the Gaussian components are used to create the codebook and quantise the audio frames into discrete symbols. The assignment of frame $\mathbf{x}_i$ to codebook index $j$ is performed using:
\vspace{-1mm}
\begin{equation}
\bar{\mathbf{x}}_t=\underset{j}{argmin} \; ||\mathbf{x}_t - \mathbf{m}_j|| \; , j \in \{1 ... V \}
\end{equation}
\vspace{-1mm}

where $\mathbf{m_j}$ is $j$th mixture component's mean vector.

To reconcile this with the LDA terminology described in Section \ref{sec:lda}, in this work each audio segment is a ``document'' and each codified audio frame is a ``word''. All the audio segments (now ``documents'') then create a whole ``collection'' or ``corpus''.

Once all the audio frames are converted to discrete ``words'', the parameters of the LDA model using $K$ domains are estimated on the $M$ audio segments from the training data using variational EM. The domain of each quantised audio segment $\bar{\mathbf{x}}$ is then given by the domain with the highest value of the posterior Dirichlet parameter $\gamma$ for that segment.
\vspace{-1mm}
\begin{equation}
	Domain(\bar{\mathbf{x}}) = \underset{j}{argmax} \; \gamma_j \, , \; j \in \{ 1 .. K\}
\end{equation}
\vspace{-3mm}

Based on the estimated parameters from the training set, Dirichlet parameters $\gamma$ can be inferred for the test set segments as well. With every segment in both train and test sets associated to a hidden domain, it is possible to perform training and/or adaptation with the usual techniques.
Acoustic models can be trained via Maximum Likelihood (ML), or domain specific models can be adapted via MAP or MLLR, in case of GMM/HMM systems.

\vspace{-1mm}
\section{Experimental setup}
\vspace{-1mm}
\label{sec:setup}
To evaluate the proposed domain discovery and adaptation method in a multi--domain and diverse ASR task, a dataset of 6 different types of data was chosen from the following sources:
\begin{itemize}
	\item Radio (RD): BBC Radio4 broadcasts on February 2009.
	\item Television (TV): Broadcasts from BBC on May 2008.
	\item Telephone speech (CT): From the Fisher corpus\footnote{All of the telephone speech data was up--sampled to 16 kHz to match the sampling rate of the rest of the data.} \cite{cieri2004fisher}.
	\item Meetings (MT): From AMI \cite{carletta2006ami} and ICSI \cite{janin2003icsi} corpora.
	\item Lectures (TK): From TedTalks \cite{USFD2014IWSLT}.
	\item Read speech (RS): From the WSJCAM0 corpus \cite{robinson1995wsjcam0}.
\end{itemize}

A subset of 10h from each domain was selected to form the training set (60h in total), and 1h from 
each domain was used for testing (6h in total). The selection of the domains aims to cover the most 
common and distinctive types of audio recordings used in ASR tasks.

Two types of acoustic features were used: First, 13 PLP features  plus 
first and second derivatives for a total of 39--dimensional feature vectors; and second, a 
65--dimensional feature vector concatenating the 39 PLP features and 26 bottleneck (PLP+BN) features 
extracted from a 4--hidden--layer DNN trained on the full 60 hours of data. 
31 adjacent frames (15 frames to the left and 15 frames to the right) of 23 dimensional log Mel filter bank features were concatenated to form a 713--dimensional super vector; Discrete Cosine Transform (DCT) was applied to this super vector to de--correlate and compress it to 368 dimensions and then fed into the neural network. The network was trained on 4,000 triphone state targets and the 26 dimensional bottleneck layer was placed before the output layer. The objective function used was frame--level cross--entropy and the optimisation was done with stochastic gradient descent and the backpropagation algorithm. DNN training was performed with the TNet toolkit \cite{vesely2010tnet} and more details can be found at \cite{liu2014using}.

For both types of features, baseline ML GMM--HMM models were trained using HTK \cite{young2006htk}
with 5--state crossword triphones and 16 gaussians per state. The language model used was based on a 50,000--word vocabulary and was trained by combination of language models from the 6 domains, with interpolation weights tuned using an independent development set.

\vspace{-1mm}
\subsection{Baseline results}
\vspace{-1mm}
Table \ref{tab:baseline} presents the baseline Word Error Rate (WER) results for the in--domain maximum--likelihood (ML) model trained with the pooled 60 hours of all domains, plus the results of ML in--domain models each trained with 10 hours of in--domain data. It also includes the MAP adapted models from the pooled model to each domain. Experiments were conducted using PLP and PLP+BN features.
The results using ML training on the limited in--domain data underperformed MAP adaptation on such data, which set MAP as a preferred setup for domain adaptation.

\vspace{-2mm}
\begin{table}[h]
	\centering
	\caption{WER (\%) of baseline models}
	\label{tab:baseline}
	\tabcolsep=0.07cm		
	\begin{tabular}{|l|l|cccccc|l|}
		\hline
		Features & Model & RS   & RD   & TK   & CT   & MT   & TV   & Total         \\ \hline\hline
		\multirow{3}{*}{PLP} & ML    & 
		17.3 & 18.4 & 34.1 & 46.6 & 44.0 & 51.1 & \textbf{36.0} \\ 
		& ML Domain & 
		16.9 & 19.1 & 35.1 & 44.4 & 44.0 & 52.9 & \textbf{36.3} \\ 	
		& MAP & 
		14.6 & 16.8 & 31.8 & 43.5 & 40.4 & 49.6 & \textbf{33.6} \\ \hline
		\multirow{3}{*}{PLP+BN} & ML  & 
		13.0 & 13.3 & 23.5 & 33.5 & 32.2 & 42.0 & \textbf{26.8} \\
		& ML Domain &
		12.6 & 14.0 & 25.0 & 34.3 & 33.2 & 44.0 & \textbf{27.9} \\
		& MAP & 
		12.1 & 12.8 & 23.1 & 32.5 & 30.6 & 41.5 & \textbf{26.2} \\ \hline
	\end{tabular}
\end{table}
\vspace{-2mm}

\section{Results}
\label{sec:results}
\vspace{-2mm}
The experiments performed aimed to evaluate two aspects of the proposed LDA modelling for unsupervised domain discovery. First, if LDA could be successfully used to find hidden domains and if these domains represented the hidden characteristics of the audio. Second, once hidden domains had been identified, if domain adaptation could be applied on them and improvements in ASR performance were achieved over the baselines.

\begin{figure}
	\centering
	\includegraphics[width=8cm]{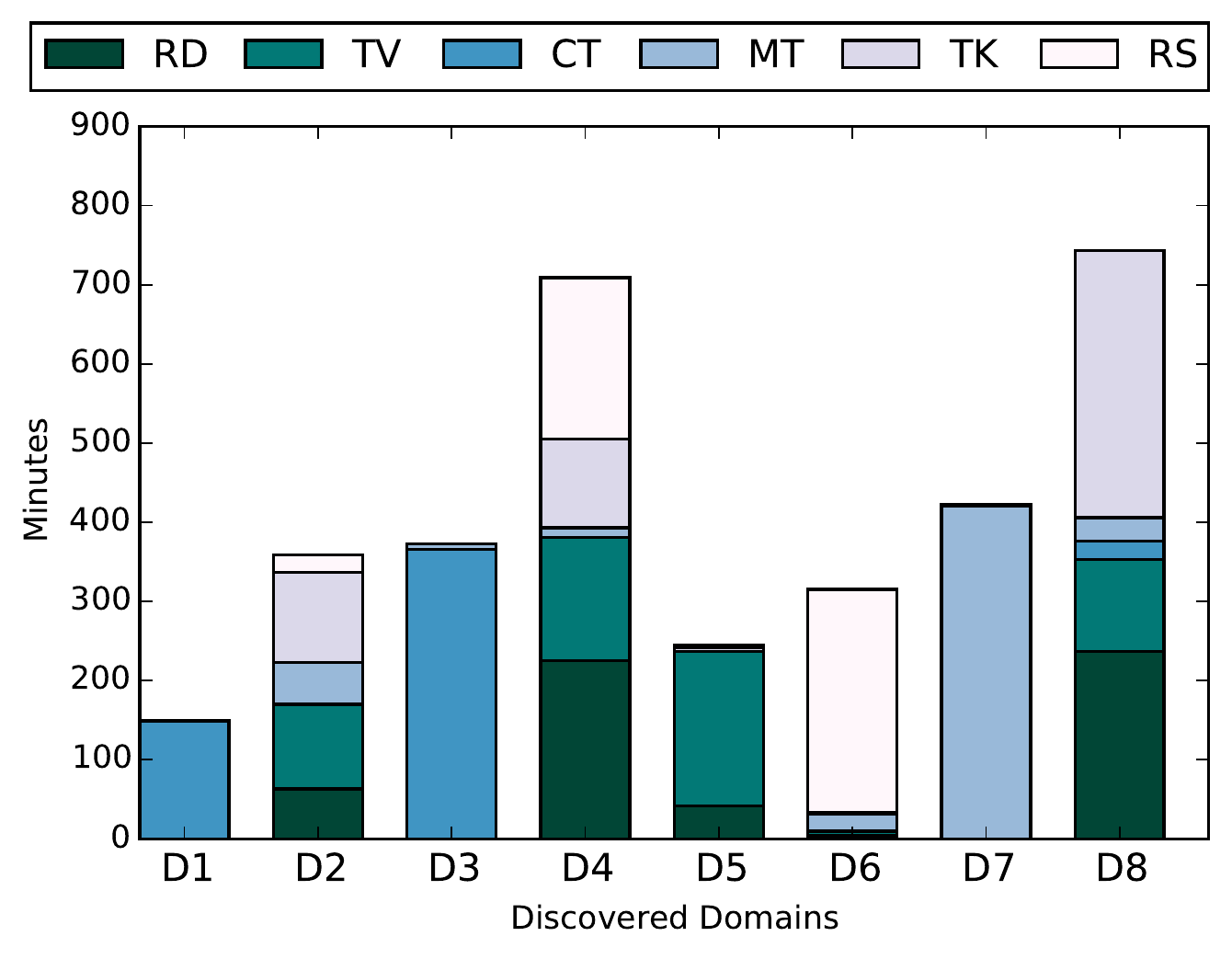}
	\caption{Amount of data for each discovered domain ($K=8$) from the labelled domains using a codebook size of 2,048}
	\label{fig:8domains-data}
\end{figure}

\subsection{Unsupervised domain discovery}

For using LDA models, as described in Section \ref{sec:lda}, two parameters had to be initially set up. First, the number of domains $K$ to be found had to be decided prior to the training. Also, since the audio frames needed to be quantised, the size of the codebook $V$ also needed to be defined. For this end, a set of experiments were conducted with different codebook sizes and number of domains. Codebooks of size 128 up to 8,192 were used and given a codebook, different LDA models with a varying number of domains from 4 to 64 were estimated \cite{hoffman2010online, rehurek_lrec} using the training data described in Section \ref{sec:setup}.

Since these identified domains were latent, there was no ground truth to verify them at this stage. An initial way of evaluating how the different latent domains behaved was by
measuring the distribution of the data, according to manual labels, which was included in each hidden domain. Figure \ref{fig:8domains-data} presents this distribution for
an acoustic codebook of size 2,048 and 8 hidden domains. From this Figure, it is possible to see how telephone speech was separated into two different hidden domains (D1 and D3), while meeting speech was mostly assigned to a unique hidden domain (D7). Other manually labelled domains, such as Radio and Television broadcasts were scattered across hidden domains (D2, D4 or D8), indicating the presence of previously unseen domains within these types of data.

Following this, KL divergence \cite{kullback1951information} was proposed as an appropriate metric to measure the consistency of the hidden topics discovered by LDA. This measured how the distributions of data in latent domains, as in Figure \ref{fig:8domains-data}, in different sets, for instance training and testing data, were different with each other:

\begin{equation}
KLD(P||Q) = \sum_{i} P(i) \ln \frac{P(i)}{Q(i)}
\end{equation}
where $P$ and $Q$ are the distributions for training and test data.
To compute the divergence, since we deal with counts in the distributions and some counts can be zero, the distributions are smoothed by discounting 3\% of the total mass and distributing it across zero counts.

\begin{figure}
	\centering
	\includegraphics[width=8cm]{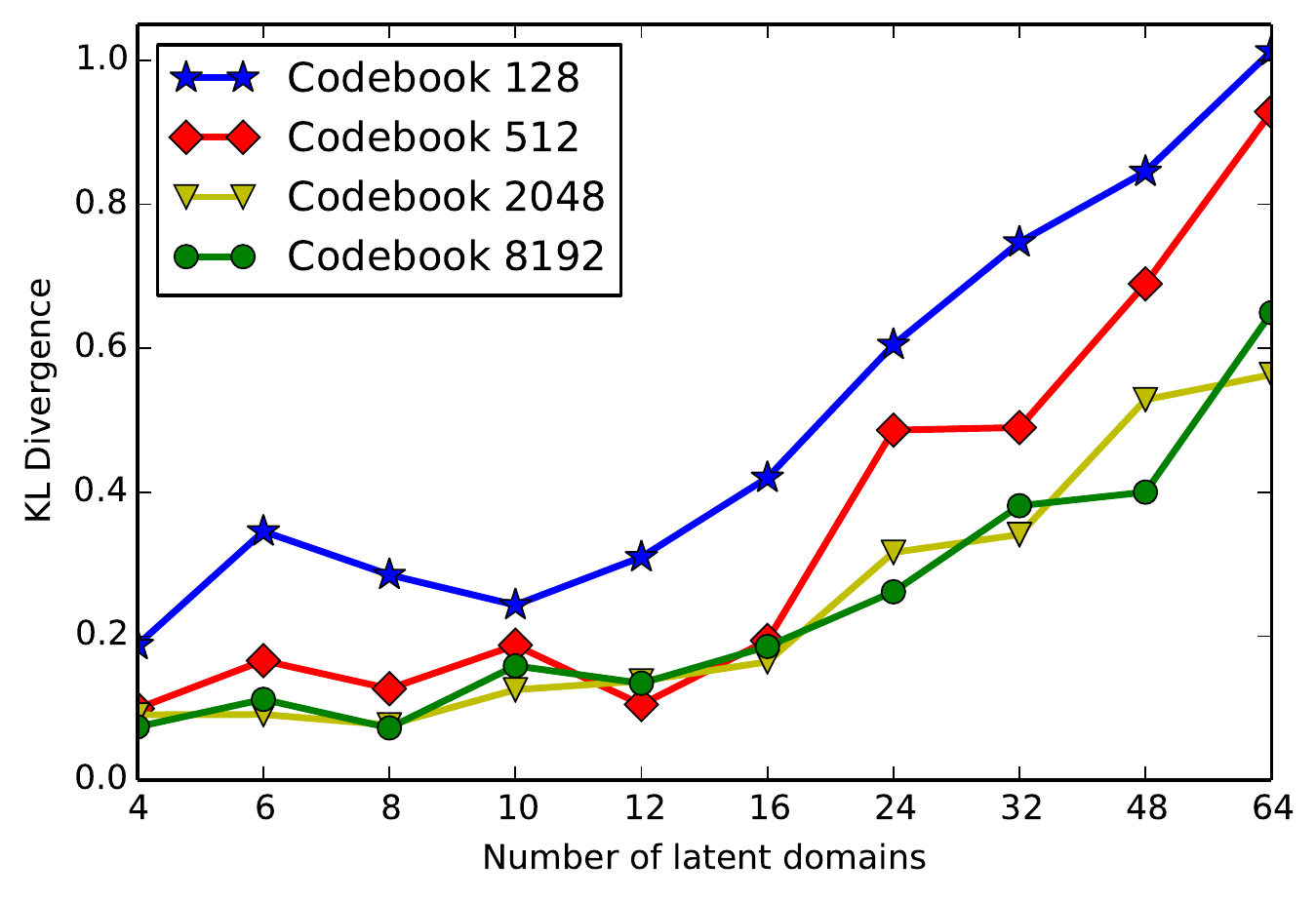}
	\caption{KL divergence of training and test set topics}
	\label{fig:divergence}
	\vspace{-3mm}
\end{figure}

Figure \ref{fig:divergence} shows the divergence values of different configurations. Low values of divergence indicated a more consistent set of hidden domains found by LDA modelling and, thus, were preferred over configurations with higher values. In terms of codebook size, codebooks of 2,048 and 8,192 symbols resulted in lower divergence. For the number of domains, increasing to more than 12 resulted an increase in divergence.

\subsection{Domain adaptation}

\begin{figure}
	\centering
	\includegraphics[width=8cm]{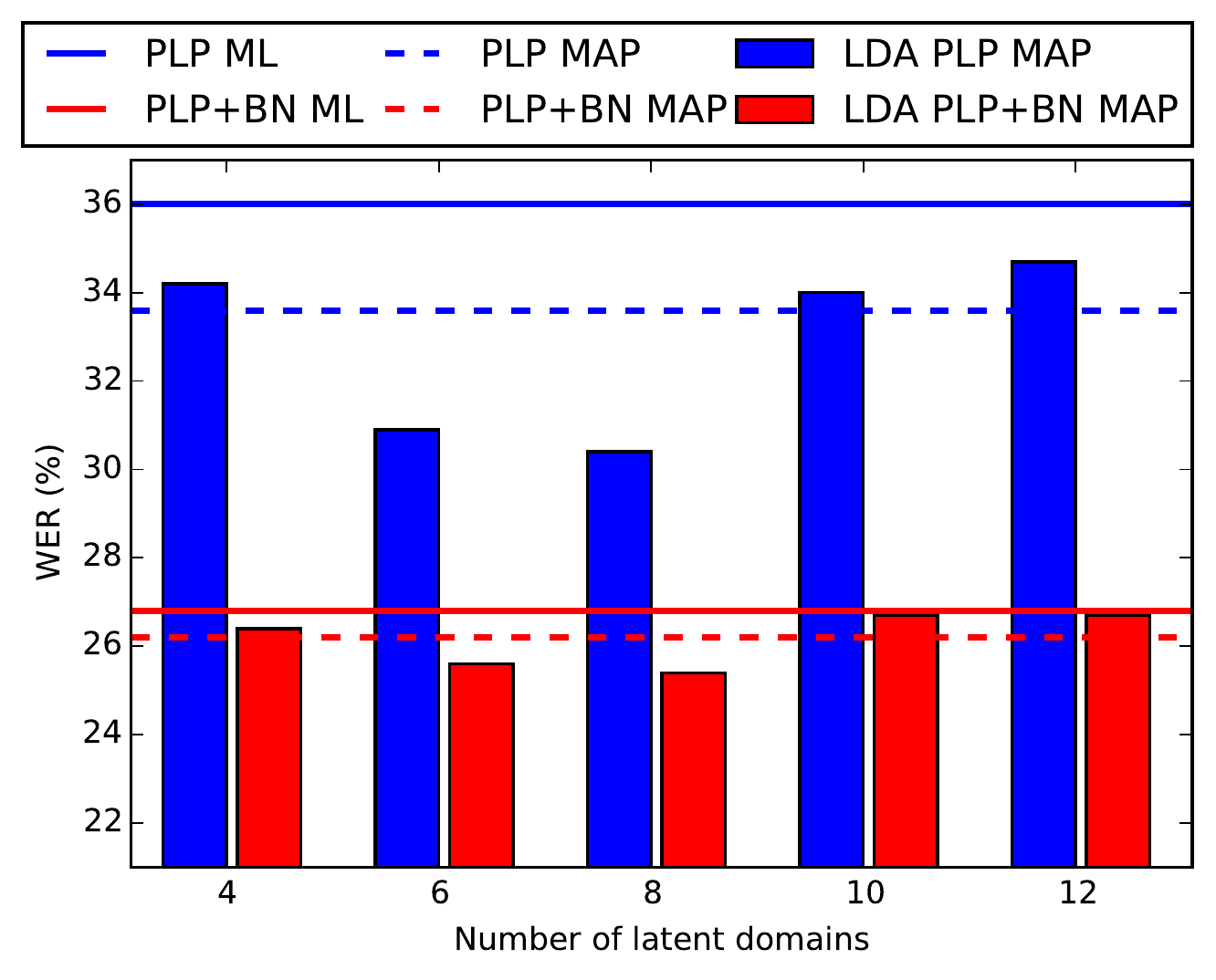}
	\caption{WER (\%) of LDA MAP adapted models with different number of topics}
	\label{fig:wer}
\end{figure}

For the evaluation of the possibilities offered by the unsupervised discovery of domains in ASR, MAP domain adaptation was performed to each of these new domains. The experiments were conducted with domains of size 4, 6, 8, 10 and 12 and a codebook of acoustic words of size 2,048. Each MAP adapted domain specific model was used to decode the corresponding speech segments in the test set that were assigned to that domain. Figure \ref{fig:wer} shows the overall WER on the test set with different number of topics using both types of features, PLP and PLP+BN. The lowest WER values, 30.4\% for PLP features and 25.4\% for PLP+BN, were achieved with 8 domains for both types of features, which was 16\% and 5\% relative improvement over their respective ML baselines. Comparing with MAP adaptation to human--labelled domains the relative WER reduction was 10\% and 3\%. The improvements in WER vanished for more than 8 hidden domains, indicating that using larger numbers of domains were not beneficial for this task.

Table \ref{tab:lda} presents the breakout of the results using 8 hidden domains across the manually labelled domains. Improvements occur across all of these domains, indicating that the LDA model can benefit all types of speech in this setup. The domains that achieved the highest gains from using LDA MAP adaptation (with PLP feature) were read speech, telephone speech and TV broadcasts, with relative WER reductions of 14\%, 12\%, 10\% respectively compared to MAP adaptation on the manually labelled domains. The lowest gain, 4\% relative, occurred on meeting speech. Similarly, with PLP+BN features telephone speech, lectures and read speech benefited the most, with relative WER reduction of 5\%, 4\% and 2\% respectively.

\vspace{-3mm}
\begin{table}[h]
	\centering
	\caption{WER (\%) of LDA MAP Models ($K=8$)}
	\label{tab:lda}
	\tabcolsep=0.07cm		
	\begin{tabular}{|l|l|cccccc|l|}
		\hline
		Features & Model & RS   & RD   & TK   & CT   & MT   & TV   & Total         \\ \hline\hline
		\multirow{2}{*}{PLP} & MAP    & 14.6 & 16.8 & 31.8 & 43.5 & 40.4 & 49.6 & \textbf{33.6}  \\
		& LDA MAP & 12.5 & 15.3 & 29.1 & 38.2 & 38.5 & 44.7 &  \textbf{30.4}  \\  \hline
		\multirow{2}{*}{PLP+BN} & MAP  & 12.1 & 12.8 & 23.1 & 32.5 & 30.6 & 41.5 & \textbf{26.2} \\
		& LDA MAP  & 11.9 & 12.8 & 22.3 & 31.1 & 31.0 & 41.0 & \textbf{25.4}  \\ \hline
	\end{tabular}
\end{table}
\vspace{-3mm}
Finally, Table \ref{tab:lda-map-new-domains} shows the WER across the hidden domains for both types of features with LDA MAP models. The most relevant feature of these domains, in terms
of WER, was that the domains of low WER (like Read speech) or high WER (like TV data) had been broken up in different hidden domains and hence, WERs across hidden domains were evenly distributed.


\vspace{-2mm}

\begin{table}[h]
	\centering
	\caption{WER (\%) of LDA MAP Models ($K=8$) across hidden domains}
	\label{tab:lda-map-new-domains}
	\tabcolsep=0.07cm		
	\begin{tabular}{|l|cccccccc|l|}
		\hline
		Features & D1   & D2   & D3   & D4   & D5   & D6 & D7 & D8   & Total         \\ \hline\hline
		PLP & 37.3 & 34.9 & 39.7 & 39.2 & 24.6 & 17.1 & 38.7 & 22.9 & \textbf{30.4} \\ \hline
		PLP+BN & 33.9 & 29.2 & 30.4 & 32.8 & 19.7 & 12.6 & 30.9 & 19.2 & \textbf{25.4} \\ \hline
	\end{tabular}
\end{table}

\vspace{-3mm}
  \section{Conclusions}
  \label{sec:conclusion}   

A novel technique based on Latent Dirichlet Allocation (LDA) has been proposed to discover latent domains in highly--diverse speech data in an un--supervised manner. The data set consisted of data from TV and radio shows, meetings, lectures, talks and telephony speech with a 60--hour training set and 6--hour test set. It was assumed that there are a set of hidden domains and each audio segment is a mixture of different properties of those hidden domains with different weights. LDA models were used to discover the latent domains and then these domains were used to perform Maximum A Posteriori (MAP) domain adaptation. Results showed relative improvement of up to 16\% over the baseline Maximum Likelihood trained models and up to 10\% over the MAP adapted models to human labelled domains with the LDA discovered domains.

The bag--of--words assumption in LDA model does not take the order of words into account. In applying LDA for image processing, there are some variants of the original LDA model, such as Spatial LDA \cite{wang_slda} which encodes spatial structure with the visual words. A temporal variant of LDA could better handle the temporal nature of speech and needs to be investigated as a future work. Also applying the current technique on bigger and/or less diverse data set needs to be verified to see what would be the new discovered domains and how they are related to domain adaptation. Newer sets of features, better targeted to describe background acoustic characteristics \cite{saz2014slt}, could also provide an improvement over PLP features, which are known to describe well phonetic and speaker information.

\vspace{-3mm}
\section{Acknowledgements}\label{sec:acknowledgements}  
This work was supported by the EPSRC Programme Grant EP/I031022/1 Natural Speech Technology (NST).
\vspace{-3mm}
\section{Data Access Statement}
The speech data used in this paper was obtained from the following sources: Fisher Corpus (LDC catalogue number LDC2004T19), ICSI Meetings corpus (LDC2004S02), WSJCAM0 (LDC95S24), AMI corpus (DOI number 10.1007/11677482\_3), TedTalks data (freely available as part of the IWSLT evaluations), BBC Radio and TV data (this data was distributed to the NST project's partners with an agreement with BBC R\&D and not publicly available yet).


The specific file lists used for training and testing in the experiments in this paper, as well as result files can be downloaded from \url{http://mini.dcs.shef.ac.uk/publications/papers/is15-doulaty2}.
\eightpt
\bibliographystyle{IEEEtran}
\bibliography{references}

\end{document}